\documentclass[conference]{IEEEtran}
\IEEEoverridecommandlockouts
\usepackage{cite}
\usepackage{amsmath,amssymb,amsfonts}
\usepackage{algorithm}
\usepackage{algpseudocode}
\usepackage{graphicx}
\usepackage{textcomp}
\usepackage{xcolor}
\usepackage{subcaption}
\usepackage{hyperref}
\usepackage{cleveref}
\usepackage{algorithm}
\usepackage{url}

\def\BibTeX{{\rm B\kern-.05em{\sc i\kern-.025em b}\kern-.08em
    T\kern-.1667em\lower.7ex\hbox{E}\kern-.125emX}}
\begin{document}

\title{Trixi the Librarian\\
}

\author{
\IEEEauthorblockN{Fabian Wieczorek}
\IEEEauthorblockA{\textit{dept. informatik, TAMS} \\
\textit{University of Hamburg}\\
Hamburg, Germany \\
fabian.wieczorek@uni-hamburg.de\\\\}
\IEEEauthorblockN{Björn Sygo}
\IEEEauthorblockA{\textit{dept. informatik, TAMS} \\
\textit{University of Hamburg}\\
Hamburg, Germany \\
bjoern.sygo@uni-hamburg.de}
\and
\IEEEauthorblockN{Shang-Ching Liu}
\IEEEauthorblockA{\textit{dept. informatik, TAMS} \\
\textit{University of Hamburg}\\
Hamburg, Germany \\
shang-ching.liu@studium.uni-hamburg.de \\\\}
\IEEEauthorblockN{Mykhailo Koshil}
\IEEEauthorblockA{\textit{dept. informatik, TAMS} \\
\textit{University of Hamburg}\\
Hamburg, Germany \\
mykhailo.koshil@studium.uni-hamburg.de}
}

\maketitle

\begin{abstract}
(Fabian) In this work, we present a three-part system that automatically sorts books on a shelf using the PR-2 platform. The paper describes a methodology to sufficiently detect and recognize books using a multistep vision pipeline based on deep learning models as well as conventional computer vision. Furthermore, the difficulties of relocating books using a bi-manual robot along with solutions based on MoveIt and BioIK are being addressed. Experiments show that the performance is overall good enough to repeatedly sort three books on a shelf. Nevertheless, further improvements are being discussed, potentially leading to a more robust book recognition and more versatile manipulation techniques.
\end{abstract}


\section{Introduction (Mykhailo)}
While the use of industrial robots is already widespread, the use of service robots remains limited\cite{sostero2020automation}. There are a few reasons for this. First, implementing the use of the service robot in the business requires not only the acquisition of the robot itself but also leads to changes in the organization like training the personnel, adapting the environment for the robot, etc. Second, while the industrial robot is used in a highly structured environment, the use case of the service robot implies working alongside humans, thus having less structure and more unseen situations. This leads to the high complexity of design, and as a result service robots often fail and require human intervention, which puts their commercial use in question. 

Therefore the main purpose of this work is rather showcasing and testing the feasibility of automating the work of a librarian on the task of manipulating the books on the shelf, rather than creating a commercially viable product. As the original plan included interaction with the visitors, this puts our system in the category of `Professional Social Service Robots` according to the \cite{lee2021service}. In this work, we aim at creating a service robot that will work in the library, so it can be deployed on the site without major changes to accommodate the robot. 

The project is based on the PR-2 platform that is available in our department Fig.~\ref{fig:PR2_WITH_HANDS}, and which is suitable for bi-manual manipulation. 
To tackle the task of book manipulation was divided into two sub-tasks: manipulation and perception. And while both of these tasks were solved to some extent, the main contribution of this work is combining them in form of a librarian robot that can operate in the library environment with as few modifications as possible.

\section{Related Works}

\subsection{Book manipulation (Björn / Mykhailo) }
There already exist different concepts on librarian robots. In example UJI librarian robot \cite{ramos2003autonomous},\cite{uji_librarian_robot}. It utilizes a single Mitsubishi PA-10 arm mounted on a mobile base. It can move around the library, locate the wanted shelf and book, and retrieve it using a specially customized two-finger gripper. Also, experiments have been made using the UJI robot equipped with a three-finger gripper to grasp books, using tactile sensors \cite{book_tilting_paper}. The motion planning for the book tilting is investigated further in \cite{9636782}. Here, the authors develop a probabilistic motion planning algorithm that allows for planning in low-dimensional sub-manifolds, created by the constraints in the planning space. The developed planner was then tested in a scenario similar to ours and using MoveIt \cite{MoveIt} framework.

A lot of approaches rely on environment modification in order to facilitate the robot's functioning, mostly to solve the navigation and object detection. For example, a robot that was designed to work in a highly structured environment, where it would pick and arrange books \cite{RFID_librarian}. It utilizes a two-finger gripper to pick the books. The robot relies heavily on landmarks in the environment, utilizing a floor with radio-frequency identification (RFID) tags to navigate and an intelligent bookshelf to locate the books.
Recent work \cite{yu2019positioning} explores the librarian scenario using the robot similar to \cite{uji_librarian_robot}, but focuses on navigation and position using QR code and binocular vision, rather than manipulation. So the book manipulation is done by using a special parallel gripper and a predefined position for placing. 

Some works focus primarily on manipulation. In \cite{9551507}, the authors investigate the use of a bi-manual setup with a suction gripper, in a setting similar to ours and train the fully connected network to predict which object to support with a non-suction gripper for the safe extraction of the selected object. The network is trained to perform in an environment similar to ours, i.e. bookshelf. Other works solve the book grasping outside of the library environment. One example would be \cite{suction_gripper}, which utilized a combination between suction and a two-finger gripper for grasping books in different configurations.

There is no readily available solution to our knowledge, that would automate book manipulation in the library environment similar to ours. 

\subsection{Perception (Shang-Ching Liu / Fabian)}
In the perception related tasks, we try to model the real-time books in the scene and furthermore matching individual book to the known book database.
Thus, we dig into each part for previous achievement.

For detection method like YOLO\cite{redmon2016you} or Fast r-cnn\cite{girshick2015fast} are two main method directions in state-of-art, the YOLO is more efficient with bounding box output and Fast r-cnn gives precise segmentation result. The evolution model of YOLO — YOLO-v5\cite{Glenn2022} have well documentation and robust pipeline and utilities such as Roboflow\cite{Roboflow} to proceed fine-tuning technique to extract the book spine, which we choose as the approach for book spine detection.

For book matching there are SIFT\cite{ng2003sift} to find the key-point of the picture, HSV (for hue, saturation, value)\cite{enwiki:1104118455} histogram to understand the color encoding, fuzzywoozy to measure the text similarity between detection text with book title in database. 

Inventory Management in a library is a tedious task that has been tried to automize in the past decade. Book spines standing on the shelf were attempted to be detected and recognized using different computer vision methods without the aid of special markers. A frequently seen approach to detect book spines is to use edge detection along with further line segment processing \cite{Tabassum2017AnAT, Nevetha2015AutomaticBS, Duan2012IdentifyingBI, Cao2019BookSR}. Often, an orthographic representation of the book spines is required. To detect the spines independently from the viewpoint, Talker et al. used a constrained active contour model allowing the spines to be non-parallel to the image axis \cite{Talker2014ViewpointindependentBS}.

The detection part is crucial to find book spine candidates, but recognizing them correctly plays an evenly important role in inventory management. While many approaches focus on text recognition to identify the book spines \cite{Tabassum2017AnAT, Nevetha2015AutomaticBS, Duan2012IdentifyingBI, Cao2019BookSR}, Fowers et al. made use of \emph{difference of gaussian} (DOG) over the YCbCr color space to extract features \cite{Fowers2010ImprovedLS}. In combination with SIFT, this approach does not depend on an OCR engine (e.g. Tesseract \cite{Smith2007AnOO}) while yielding robust performance.

Comparing the results of the mentioned work meaningfully is hard as no standardized way in the field of book spine recognition exists. However, in the domain of scene text recognition (STR), which can be utilized for text-based book spine recognition, a framework was developed by Baek et al. to allow comparison of different model architectures \cite{baek2019STRcomparisons}. Since deep learning methods have generally not been widely applied in book spine recognition, this work tries to incorporate such an STR model to perform text matching. 

\section{System Overview (Shang-Ching Liu)}
\begin{figure}[ht]
    \centering
\includegraphics[width=8cm]{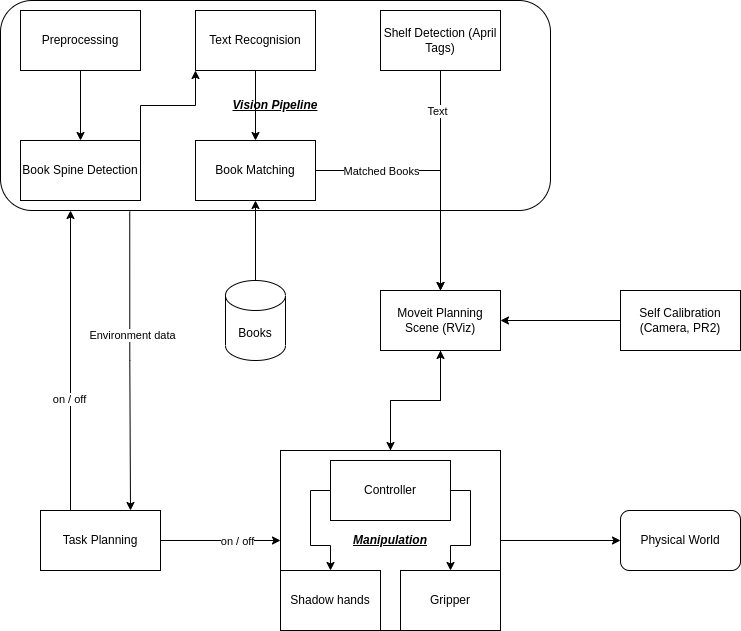}
    \caption{System Overview}
    \label{fig:system_overview}
\end{figure}

\begin{figure}[ht]
    \centering
\includegraphics[width=8cm]{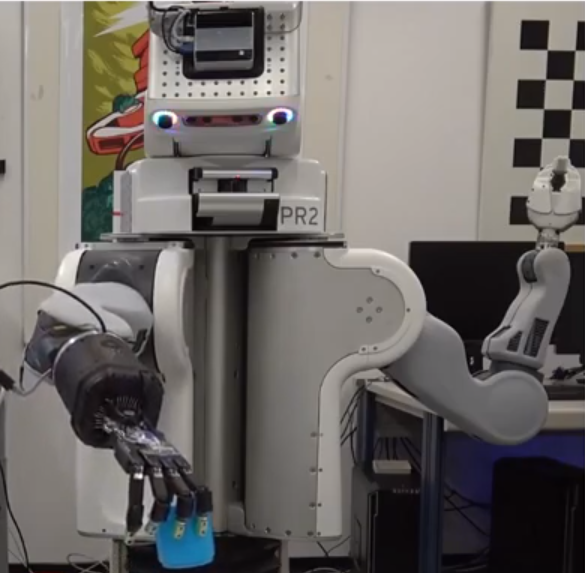}
    \caption{PR-2 with shadow hands on the right-hand side and normal two finger grippers on the left-hand side.}
    \label{fig:PR2_WITH_HANDS}
\end{figure}

The central system can be separated into three parts, including Manipulation, vision pipeline, and task planning, as shown in figure \ref{fig:system_overview}. Task planning module controlling vision pipeline module and Manipulation module. The Vision pipeline takes the RGBD camera (Azure Kinect) scene as input and matches the books in the scene to the books database, and finally creates a MoveIt Planning Scene in visualization. The Manipulation has a controller to control two hands of the Robot (PR-2). One is combined with Shadow hands, and another is combined with a two fingers gripper, as shown in the figure \ref{fig:PR2_WITH_HANDS}j.

\section{Perception (Fabian / Shang-Ching Liu)}

\subsection{Preprocessing (Fabian)}
\begin{figure}[ht]
    \centering
\includegraphics[width=.49\textwidth]{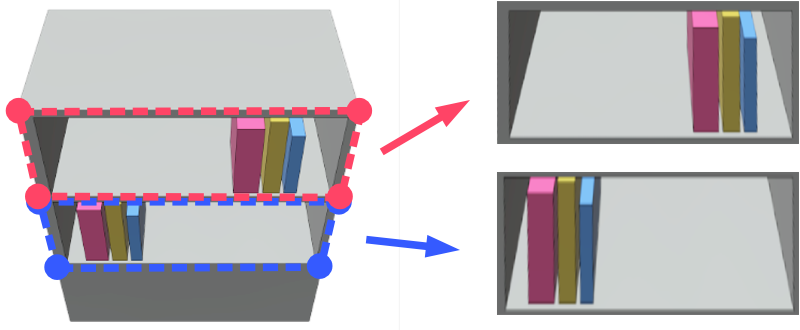}
    \caption{Visualization of the perspective transformation. The corners of each shelf level (red/blue dots) are used as anchors for the transformation. The result is an image of each shelf level having the book spines aligned with the image axis (right).  }
    \label{fig:perspective}
\end{figure}

The camera of the PR-2 is located on top of its head. This will create a perspective projection of the shelf (\cref{fig:perspective}, left; \cref{fig:inverse_perspective}, right) making book spine detection more challenging since the edges tend to be not aligned with the image axis. To automatically mitigate this problem without rearranging the hardware, a perspective transformation is applied to the image twice as shown in \cref{fig:perspective}. For each shelf level, the corners (red/blue dots) are determined using the AprilTags known pose along with offsets matching the shelfs dimensions. The 3D points are then projected onto the image to be used as anchor points for the transformation. The results are two images with the book spine edges being aligned with the image axes (\cref{fig:perspective}, right). 

To project back from the corrected images to the original image, the inverse projection matrix is also computed and will be used later on.

\subsection{Book Matching}
The section below describing the approach we have used to tackle down the book matching tasks.

\subsubsection{Book database (Shang-Ching Liu)}
We collect the following information from each books as shown in table \ref{table:database_explanation}.

\begin{table}[ht]
\begin{center}
\begin{tabular}{||c c||} 
 \hline
 column & note \\ [0.5ex] 
 \hline\hline
 id & incremental number of unique id   \\ 
 \hline
 title & book title \\
 \hline
 height & book height  \\
 \hline
 width & book width \\
 \hline
 depth & book depth \\
 \hline
 author & book author \\
 \hline
 cover\_type & hard/soft cover of book \\
 \hline
 count & Number of specific book \\ [1ex] 
 \hline
\end{tabular}
\end{center}
\caption{\label{table:database_explanation} Books Database}
\end{table}

\subsubsection{Book Spine Recognition (Shang-Ching Liu)}
\begin{figure}
    \centering
    \includegraphics[width=8cm]{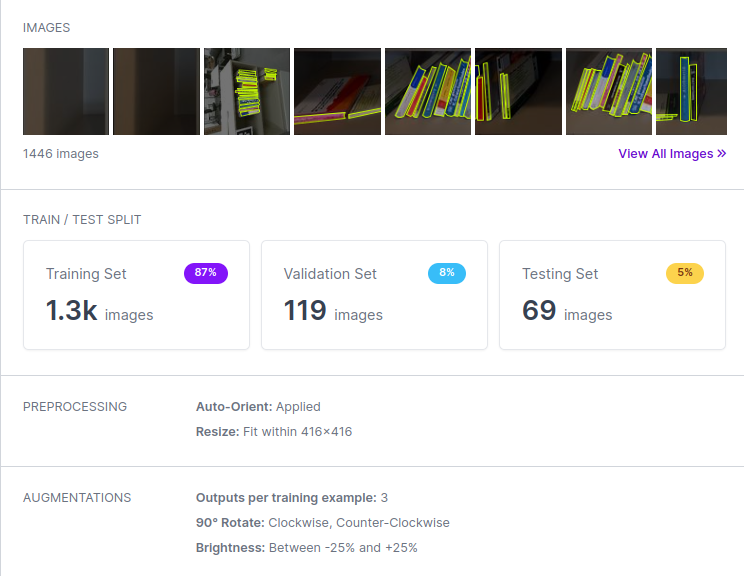}
    \caption{The book spine dataset, we have already split it to training, validation, and testing set. You can access by link\cite{book_spine_dataset}}.
    \label{fig:book_spine_dataset}
\end{figure}

Turning now to book recognization. The book spine is intuitively the hint from the front side perspective. In the beginning, we tried object detection for the book, which is one of the initial labels in the COCO dataset\cite{lin2014microsoft}. However, it failed to recognize the book spine and had a critical issue for recognizing multiple books spine together as one book spine. So that we start to collect our own dataset for book spine data to fine-tuning the model. We have created the dataset\cite{book_spine_dataset} which labeled book spine for 611 images and done the data augmentation as describe in figure \ref{fig:book_spine_dataset}.   

\subsubsection{Text Detection (Fabian)}
\begin{figure}[ht]
    \centering
    \includegraphics[width=.49\textwidth]{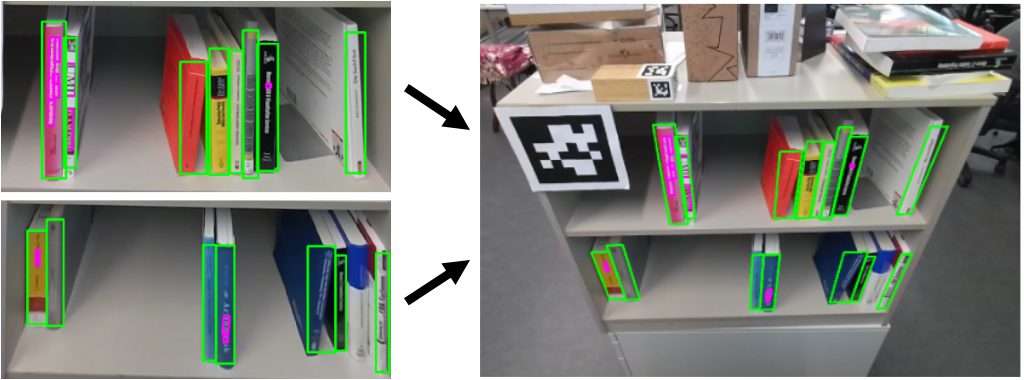}
    \caption{Visualization of reversing the perspective transformation along with the bounding boxes of detected spines (green) and text (magenta). The vertices of the polygons were transformed back so they line up with the source image (right). }
    \label{fig:inverse_perspective}
\end{figure}

In order to make use of the scene text recognition model proposed in \cite{baek2019STRcomparisons}, the image needs to be cropped so that only the text of interest is visible. To get the bounding boxes of each text segment, the CRAFT model \cite{baek2019character} is being used. Running the model on the two preprocessed input images results in the desired bounding boxes for each detected text pieces in the form of polygons (see \cref{fig:inverse_perspective}, magenta boxes). However, the vertices of the polygons are still aligned with the transformed images and not the source image. To reverse the perspective transformation applied during preprocessing, each point is transformed using the inverse matrix (\cref{fig:inverse_perspective}).

\subsubsection{Text Recognition (Fabian / Shang-Ching Liu)}

\paragraph{DTRB Model (Fabian)}
To recognize the text coming from the text detection, the proposed model from the deep-text-recognition-benchmark (DTRB) \cite{baek2019STRcomparisons} is used. The authors compared different combinations of model components against each other and existing STR models. They showed that a combination of thin-plate splines (TPS) \cite{DBLP:journals/corr/JaderbergSZK15}, ResNet\cite{He2016DeepRL}, Bidirectional LSTM (BiLSTM) \cite{Graves2005FramewisePC}, and attention-based sequence prediction (ATTN) \cite{Cheng2017FocusingAT} shows the best performance while also taking the most time to compute. With this combination, each text detection is turned into actual text and can be used as a feature for the final classification.

\paragraph{Google Vision API (Shang-Ching Liu)}
In addition, we integrate Google Vision API to conquer the difficulty of text recognition issues due to size, light, font, or multi-language in the scene. The pros would be
Track changes is off
Everyone
Track changes for everyone
You
Track changes for You
fabian.wiecz
Track changes for fabian.wiecz
goerner
Track changes for goerner
6sygo
Track changes for 6sygo
mishamail7
Track changes for mishamail7
Guests
Track changes for guests
Added iz panel as figure \ref{fig:result_rviz}… (show all)
Sep 30, 2022 10:07 PM • You
Current file
Overview
Trixi the Librarian
Fabian Wieczorek
dept. informatik, TAMS
University of Hamburg
Hamburg, Germany
fabian.wieczorek@uni-hamburg.de
 generalized text recognition and able to generate quite a robust result. The drawback, in reality, is that most of the time, the camera still fails to give a clear enough image to do the text recognition. On the other hand, the Vision API from Google is limited to free usage 1000 times a month.

\subsubsection{HSV histograms of book spines (Fabian)}
While text recognition can serve as a distinct feature for matching, it suffers as soon as the text becomes illegible e.g. due to being too small. To make the classification more robust, the histogram over the HSV color space is used as another feature similar to \cite{Fowers2010ImprovedLS}. For each detected book spine, the histogram over each channel is computed with a resolution of 20 bins. Then, a score on how much a spine detection resembles to a book from the database is determined using the cosine similarity of both histograms.

\subsubsection{SIFT Matching (Shang-Ching Liu)}
\begin{figure}[ht]
    \centering
    \includegraphics[width=4cm]{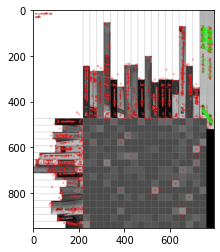}
    \caption{Successful to match the book but not for the vertical one, on the left side is all the book spine in the databases in vertical and horizontal view.}
    \label{fig:sift_successful}
\end{figure}

\begin{figure}[ht]
    \centering
    \includegraphics[width=4cm]{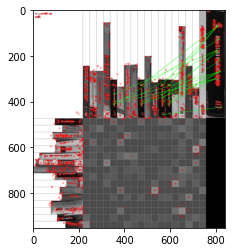}
    \caption{Failed to match the book with multiple similar book in the database, on the left side is all the book spine in the databases in vertical and horizontal view.}
    \label{fig:sift_failed}
\end{figure}

On the other hand, we try out the pure SIFT Matching for book matching. It works pretty well on some examples, as shown in figure \ref{fig:sift_successful}, but some similar books are hard to tackle, shown in figure \ref{fig:sift_failed}. Thus, we do not use SIFT as our implementation.

\subsubsection{Final Classification (Fabian)}
\label{sec:final_class}
\begin{figure}[ht]
    \centering
    \includegraphics[width=.4\textwidth]{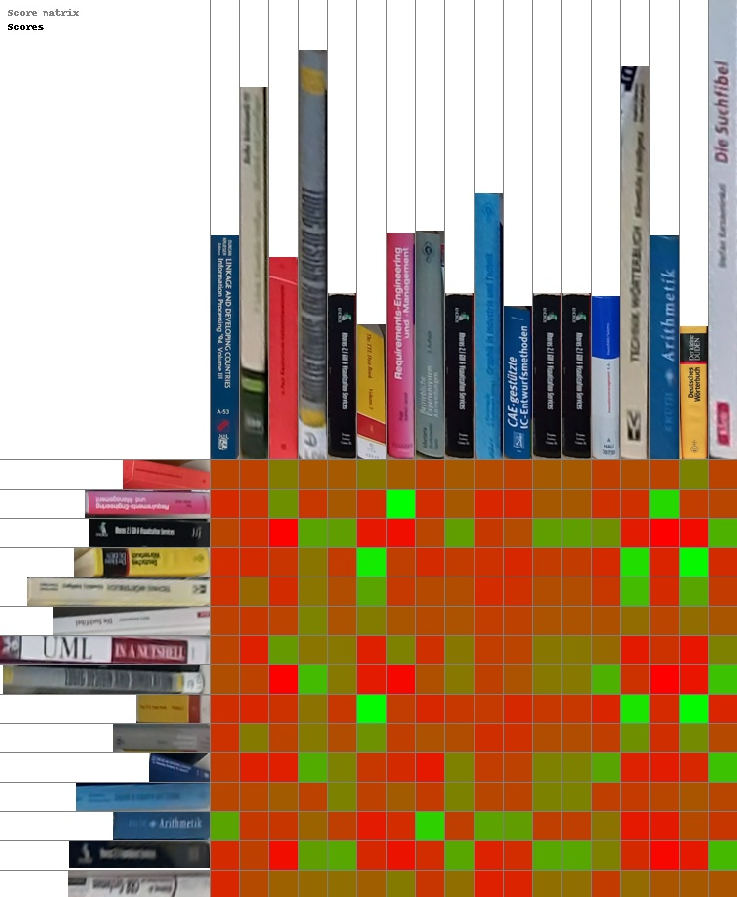}
    \caption{The final score of each detected book (rows) against the ones registered in the database (columns). Green means higher score indicating a higher similarity. }
    \label{fig:score_matrix}
\end{figure}

Since neither the text-based matching nor the SIFT matching showed sufficient performance, the final score is only computed using the HSV-histogram feature with the value-channel being ignored as well. The score was empirically defined as $x_{score} = sim_{hue} + sim_{sat} * 0.2$. \Cref{fig:score_matrix} shows each score from a set of detections against each book from the database.

To determine which detection is assigned which book ID from the database, \cref{alg:assign_id} is being used with the score matrix as input.

\begin{algorithm}
\caption{Assign book ID to each detection}
\begin{algorithmic}
\While{rows left}
\State $row, col \gets \text{findHighestScore}()$
\State assign ID from book at $col$ to detection at $row$
\State removeRow($row$)
\State removeColumn($col$)
\EndWhile
\end{algorithmic}
\label{alg:assign_id}
\end{algorithm}

This ensures that each detection is assigned an ID while each ID is only being assigned once. Furthermore, the score of each book associated with the assigned ID defines the confidence of this book recognition. 

\subsection{Physical World Simulation (Fabian)}
\begin{figure}
\begin{subfigure}[t]{.15\textwidth}
\centering
\includegraphics[width=\linewidth]{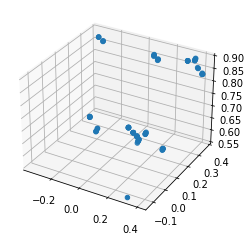}
\subcaption{140 Detections}  
\end{subfigure}
\hfill
\begin{subfigure}[t]{.15\textwidth}
\centering
\includegraphics[width=\linewidth]{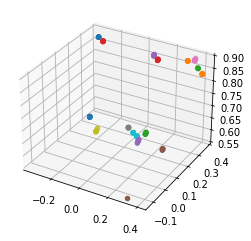}
\subcaption{After clustering}
\end{subfigure}
\hfill
\begin{subfigure}[t]{.15\textwidth}
\centering
\includegraphics[width=\linewidth]{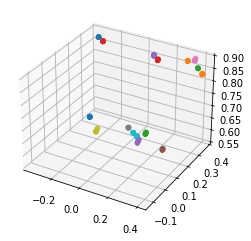}
\subcaption{After pruning}  
\end{subfigure}

\caption{Visualization of book clustering and pruning.}
\label{fig:cluster}
\end{figure}

To make the vision less prone to noise, the final belief about the environment is calculated with $10$ sequential observations. The environment consists of a set of recognized books including their pose similar to one observation. After accumulating the book sets, it can neither be assumed that each observation contains the same amount of books nor that the same books were recognized at the same position. To decide which book recognitions belong to the same book instance, the recognitions are spatially clustered using the position and the $k$-means clustering algorithm (\cref{fig:cluster}). 

The parameter $k$ is chosen to be the maximum number of book recognitions in one observation which ensures a sufficient number of centroids when clustering (\cref{fig:cluster}, b). Since miss detections should be excluded from the environment, clusters that have less than $4$ recognitions are being pruned (\cref{fig:cluster}, c). 

For each cluster, the position can be determined by averaging the positions from the associated book recognitions. To assign an ID, \cref{alg:assign_id} is used again, except the rows are substituted with clusters rather than book recognitions and the score for each book is computed as the number of ID occurrences inside the respective cluster. Furthermore, a confidence $c$ for each cluster is defined as 

\begin{equation*}
    c = \frac{\sum_{i=0}^{n} \text{conf}(b_i)}{n}
\end{equation*}

where $b$ is the list of book recognitions and $n$ is the size of the respective cluster. The function \emph{conf} returns the confidence for the given book recognition computed in \cref{sec:final_class}. With each cluster having a book ID, confidence, and a pose, the environment is complete and is then used for task planning and to create the collision objects in the planning scene. 

\subsubsection{RViz Panel (Fabian)}
\begin{figure}[ht]
    \centering
    \includegraphics[width=.49\textwidth]{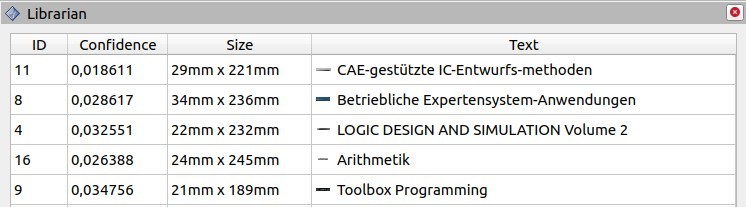}
    \caption{The RViz panel can be used to monitor the environment as it was percieved. }
    \label{fig:rviz_panel}
\end{figure}

To monitor the perceived environment described in the previous subsection, a special RViz panel is used. The panel titled \emph{Librarian} displays each book recognition as a row of a table showing the book ID, confidence, estimated dimension, and title from the database (see \cref{fig:rviz_panel}).

\section{Task Planning (Mykhailo)}

A separate task planning module is developed to process the environment state received from the perception module and issue commands to the manipulation module.

The backbone of the planning module is ROS\cite{ros}, and all the communications between the modules are done in form of the ROS messages. 

As can be seen in Fig.~\ref{fig:system_overview}, the task planning module receives a world state representation from the perception module creates a plan, and sends it for execution to the manipulation module. In order to facilitate the perception module, the task planning module issues a command to stop or start the perception. This is done to avoid possible in-scene interference caused by the grippers executing motions. 

 The world state consists of the set of the book entities with the information regarding the pose, and properties retrieved from the database. The set of possible actions is:
\begin{enumerate}
    \item perception start/stop
    \item "pick+place", pick the specified book and place it against a vertical object in the specified place and from the specified direction. 
\end{enumerate} 
We use a descriptive model of the world \cite{ghallab_nau_traverso_2016}, that is we assume that outcome of the action is determined. The desired behavior can be input by sending a message to the module with the specified command. Currently, the implemented behavior is to arrange book by the desired property on the lower shelf starting from the left corner.

Due to the limited set of action  (a book can be picked only when standing vertical) and strong assumptions on the environment (finite static environment, no explicit time, no concurrency, determinism of action success), as well as almost no precondition on the allowed actions, we have resorted to the use of the procedurally generated plans: the books on the scene are sorted regarding the desired criteria, and then moved to the lower shelf in the sorted order. This can be justified as: the cost of placing a book on the lower shelf is equal for all books (no search needed); all books should be sorted eventually; there is no action that allows for the fallback recovery. 

For future work, we plan on expanding the planning module to include more complex behavior such as recovery from the fallback scenarios and adding new desired behaviors. This can be done in the current framework by extending the state space and removing the assumption that books are only standing, adding manipulation for new states of the book. This will allow us to use {\it integration of task and motion planning} like \cite{6906924} or \cite{7139022} which are suitable for our hierarchical architecture.

\section{Manipulation (Björn)}
The book manipulation process is divided into two main procedures. First is grasping and extracting the book from the shelf. Second is placing the book into a designated position in the shelf again. The platform used is our custom fit PR-2, which has both a two finger gripper and a shadow hand on its two arms. For controlling the robot and planning, we utilized the MoveIt \cite{MoveIt} framework as well as the BioIK solver \cite{BioIK}. Specifically, we use the OMPL planner for planning the motions. For motions, which require one of the grippers to a position further away, we don't just give a pose to the OMPL planner, but instead get a joint configuration for the corresponding arm from the BioIK solver and then let the OMPL planner move to that configuration. This has proven to be more reliable to result in configurations, which can execute the following steps, through the use of more specific constraints, that can be given to the BioIK solver.

\subsection{Grasping and extracting}
The grasping process is implemented as a bi-manual procedure utilizing both arms of the PR-2. The bi-manual approach for grasping and extracting the book was only our second plan. First, we wanted to expand on the approach in \cite{book_tilting_paper} and only utilize the shadow hand. During experimentation, we wanted the shadow hand to hold the book in the tilted position with its index finger and use the other fingers to grasp the book. However, due to the limited workspace of the shadow hand, we didn't manage to manually find a position, which allowed the planning, both FK and IK planners, to find a path to grasp the book. With additional quantitative testing, we also came to the conclusion, that the workspace of the shadow hand might not allow such and operation. Another option would be to use reinforcement learning to find such a position, but with doubt, that such a position even exists, and our limited time schedule, we decided not to invest more time into it and decided on a bi-manual approach instead.

Now, the shadow hand is first used to bring the book into as position, which allows the other gripper to grasp the book around its spine. Since usually this isn't possible from the start because the book spines align, we implemented the previously mentioned process. It takes one finger of the shadow hand, places it on top of the book and starts tilting it by pulling on it. We settled on an approach, where the finger of the shadow hand is stretched out, while the other fingers are curled up. Originally, we had the finger in a hook like pose, so it could exert force more easily on the book. However, this approach ran into problems with the limited space between the book and the shelf above, making it necessary to switch to the stretched out finger.

\begin{figure}
\begin{subfigure}[t]{.22\textwidth}
\centering
\includegraphics[width=\linewidth]{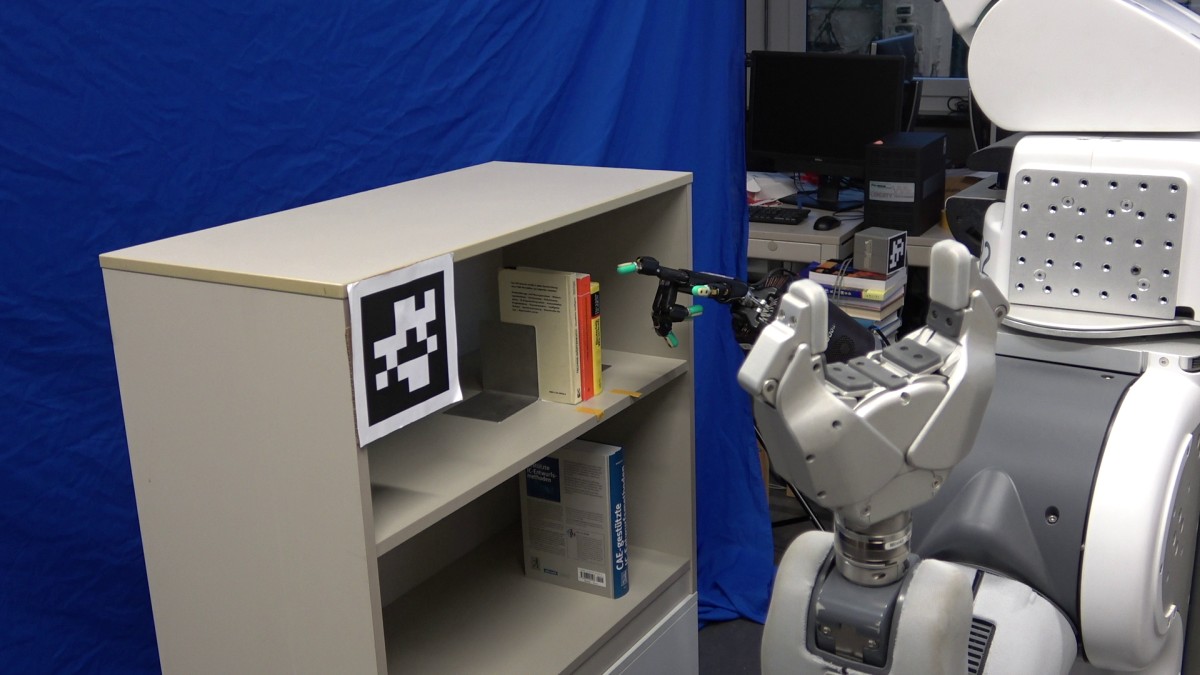}
\subcaption{Moving before the book to prepare tilting.}  
\end{subfigure}
\hfill
\begin{subfigure}[t]{.22\textwidth}
\centering
\includegraphics[width=\linewidth]{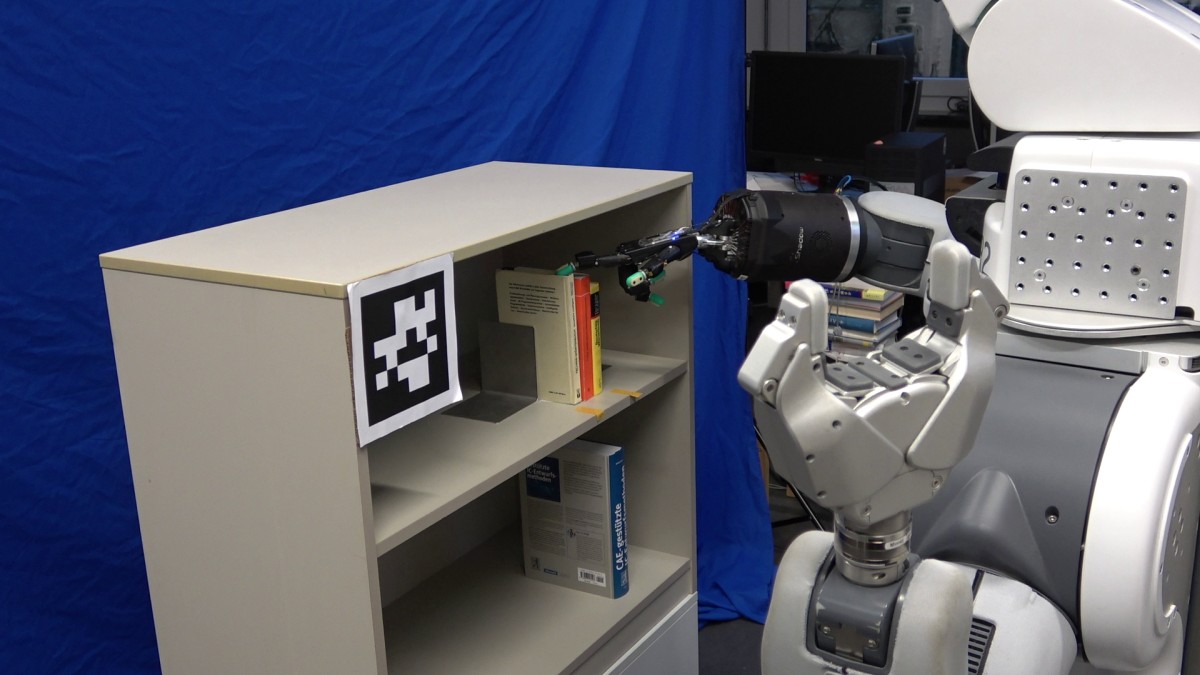}
\subcaption{Moving the finger on the book to make contact for tilting.}  
\end{subfigure}
\hfill
\begin{subfigure}[t]{.22\textwidth}
\centering
\includegraphics[width=\linewidth]{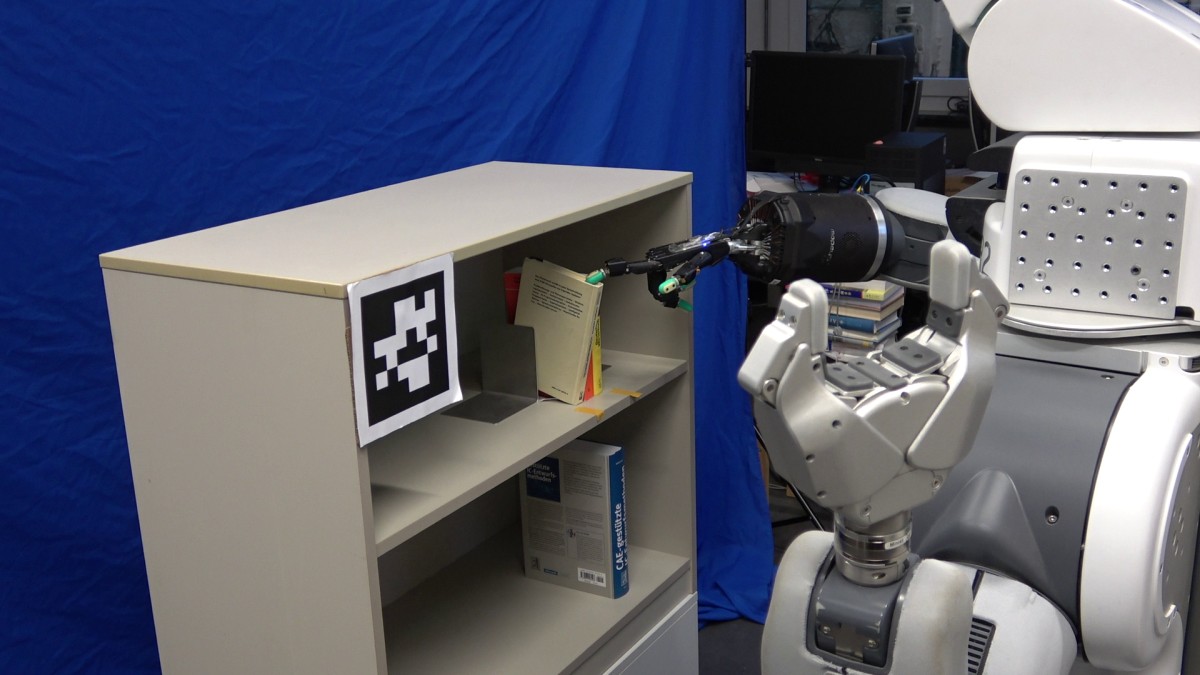}
\subcaption{Pulling back the shadow hand to tilt the book.}  
\end{subfigure}
\hfill
\begin{subfigure}[t]{.22\textwidth}
\centering
\includegraphics[width=\linewidth]{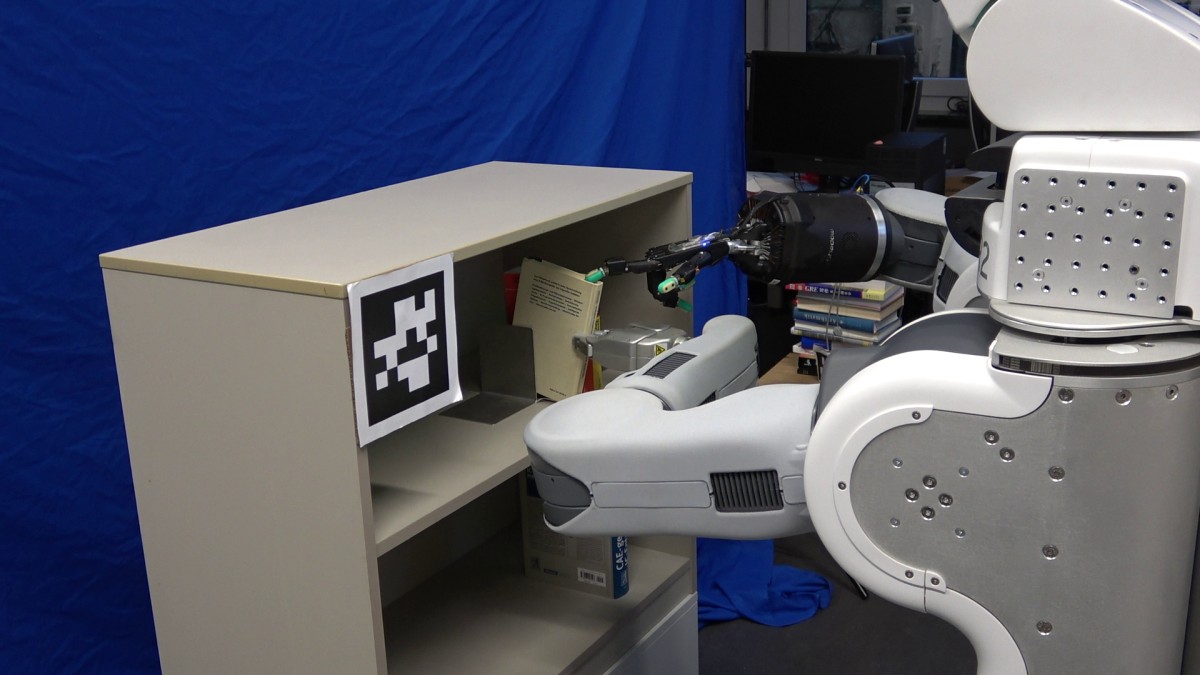}
\subcaption{Use the gripper to grasp the book while it is tilted by the shadow hand.}  
\end{subfigure}
\hfill
\begin{subfigure}[t]{.22\textwidth}
\centering
\includegraphics[width=\linewidth]{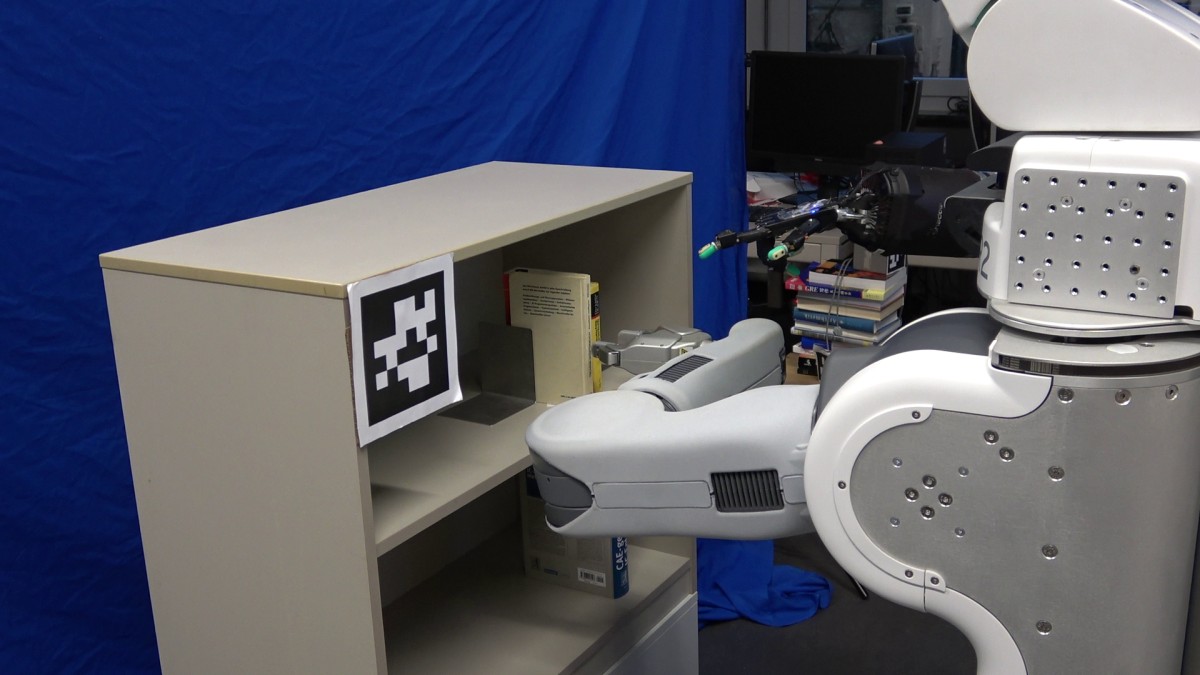}
\subcaption{Gripper after pulling the book out a bit and releasing it.}  
\end{subfigure}
\hfill
\begin{subfigure}[t]{.22\textwidth}
\centering
\includegraphics[width=\linewidth]{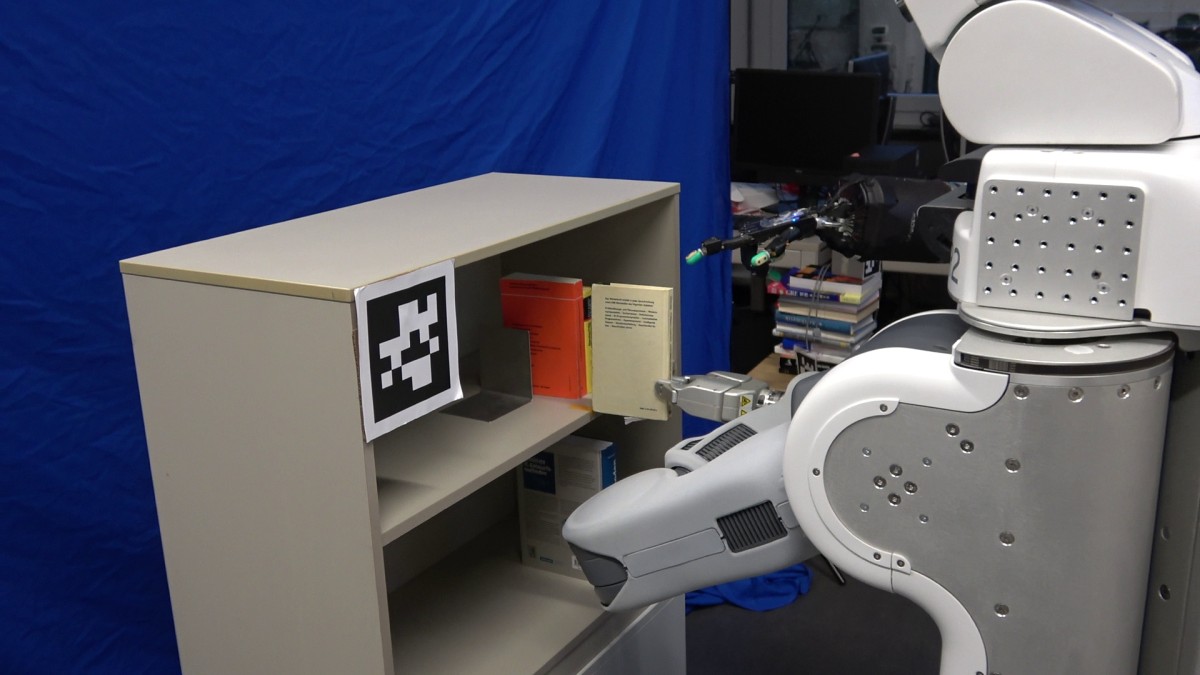}
\subcaption{Pulled out book after the gripper has grasped it again.}  
\end{subfigure}
\caption{The process of extracting the book.}
\label{fig:grasping_process}
\end{figure}

This tilting allows the gripper then to grasp the book on the newly exposed part of the book and later extract it. For this procedure to work, we need a few assumptions for the books to be true. First, their book spines need to align with each other. Second, the books need to stand upright in the shelf. Both these assumptions seem reasonable in a library environment. Further, the books are limited by weight constraints of the robot, and they need to be able to stand upright on their own consistently.
After the gripper has grasped the book, the shadow hand then removes itself from the book and the gripper extracts it. It does so by first pulling the book pack further and then letting it go to fall back. This process lets the book fall back in an upright position again with a still exposed spine, allowing the gripper to again grasp the book, but with a more stable grasp, having a larger exposed area for grasping. Then the book gets pulled out from its position. The full process can be seen in \ref{fig:grasping_process}.

We chose to use the two general purpose manipulators already mounted on our PR-2, instead of building a custom gripper, suited for book manipulation, as in \cite{uji_librarian_robot} or a more specialized gripper, for example a suction gripper \cite{suction_gripper}. Our reason for this is, that in the future, a librarian robot should also be able to perform other tasks than just manipulating books, for example interact with customers or other objects in the library. Therefore, a specialized gripper for just the task of manipulating books could hinder the robot to fulfill these tasks, which is why we wanted to try to allow the robot to complete this task without special customization.

\subsection{Placing}
\begin{figure}
\begin{subfigure}[t]{.2\textwidth}
\centering
\includegraphics[angle=180, width=\linewidth]{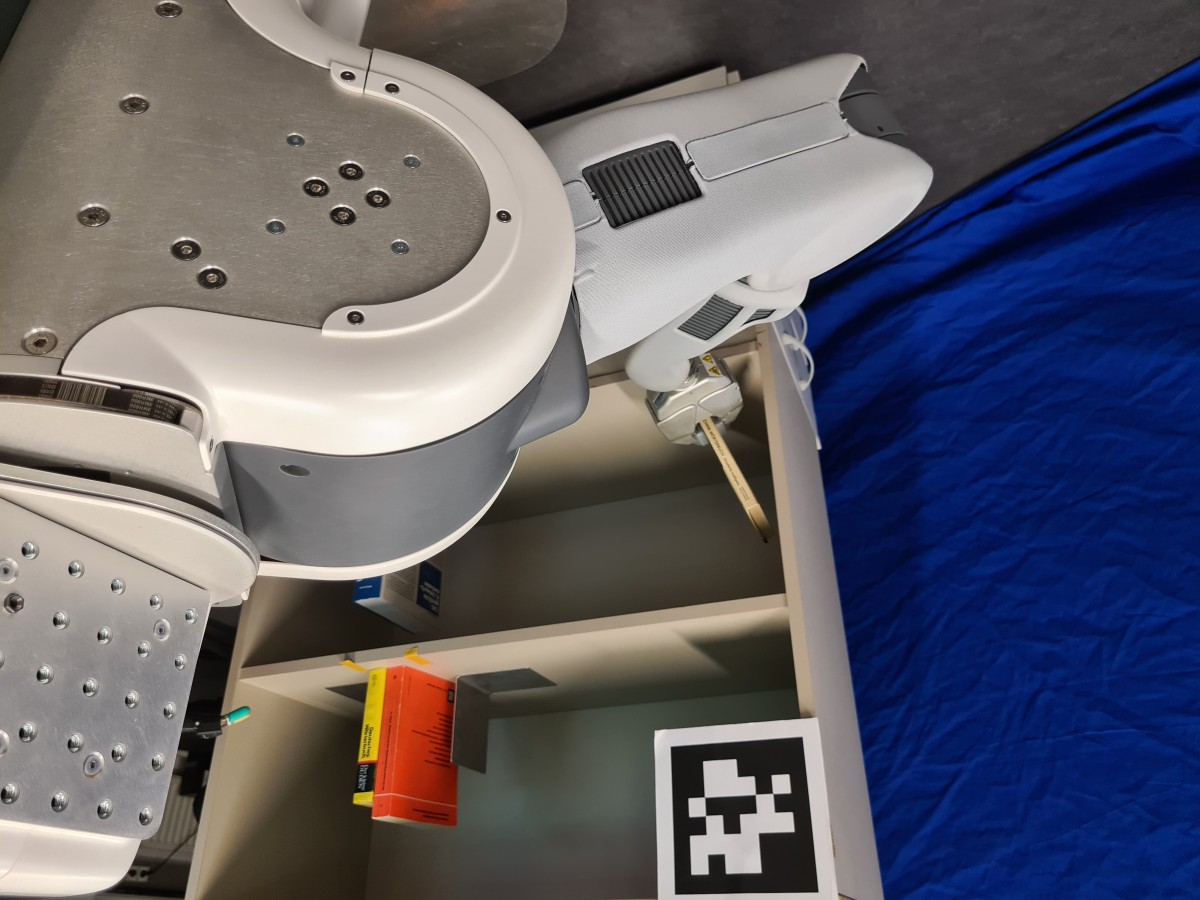}
\subcaption{Moving the book into its initial position in the shelf before placing.}  
\end{subfigure}
\hfill
\begin{subfigure}[t]{.2\textwidth}
\centering
\includegraphics[angle=180, width=\linewidth]{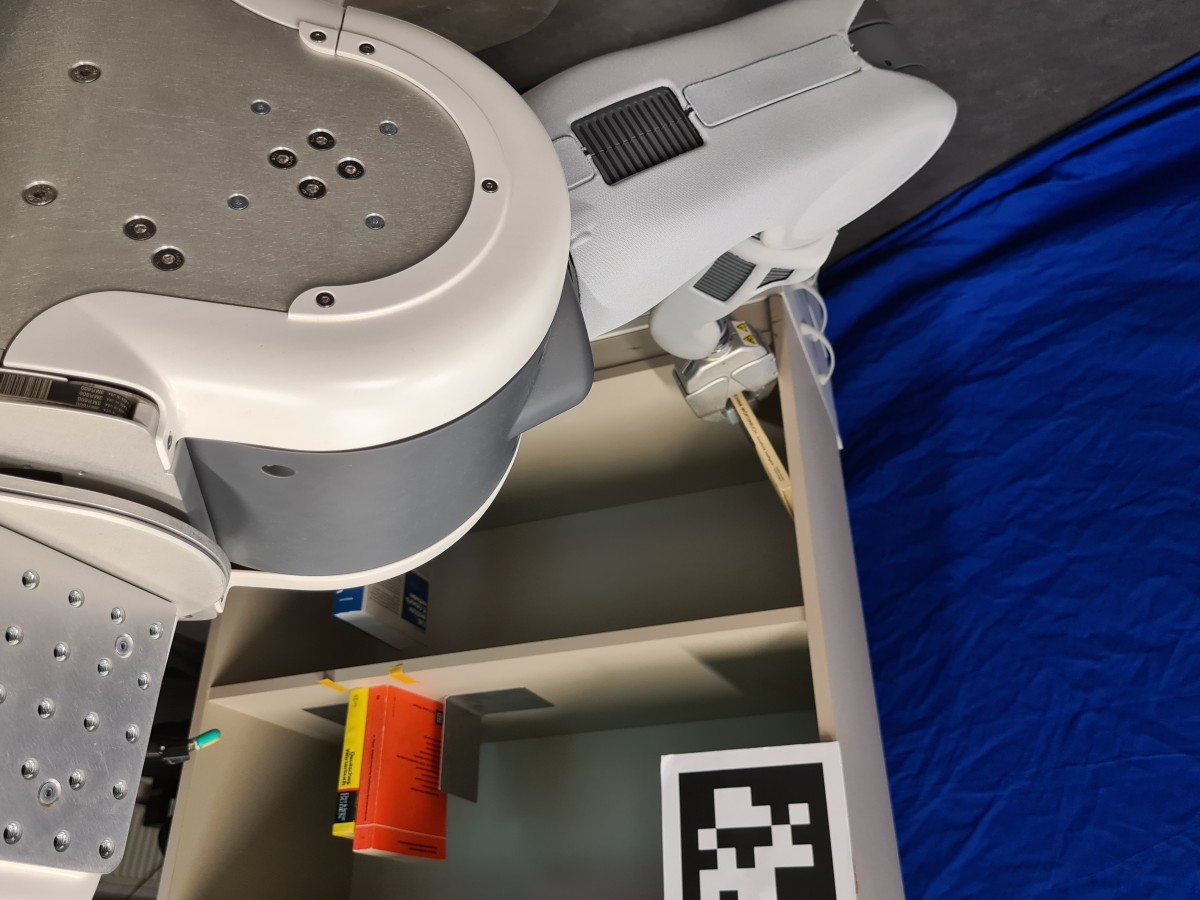}
\subcaption{Leaning the book against the shelf to the left.}  
\end{subfigure}
\begin{subfigure}[t]{.2\textwidth}
\centering
\includegraphics[angle=180, width=\linewidth]{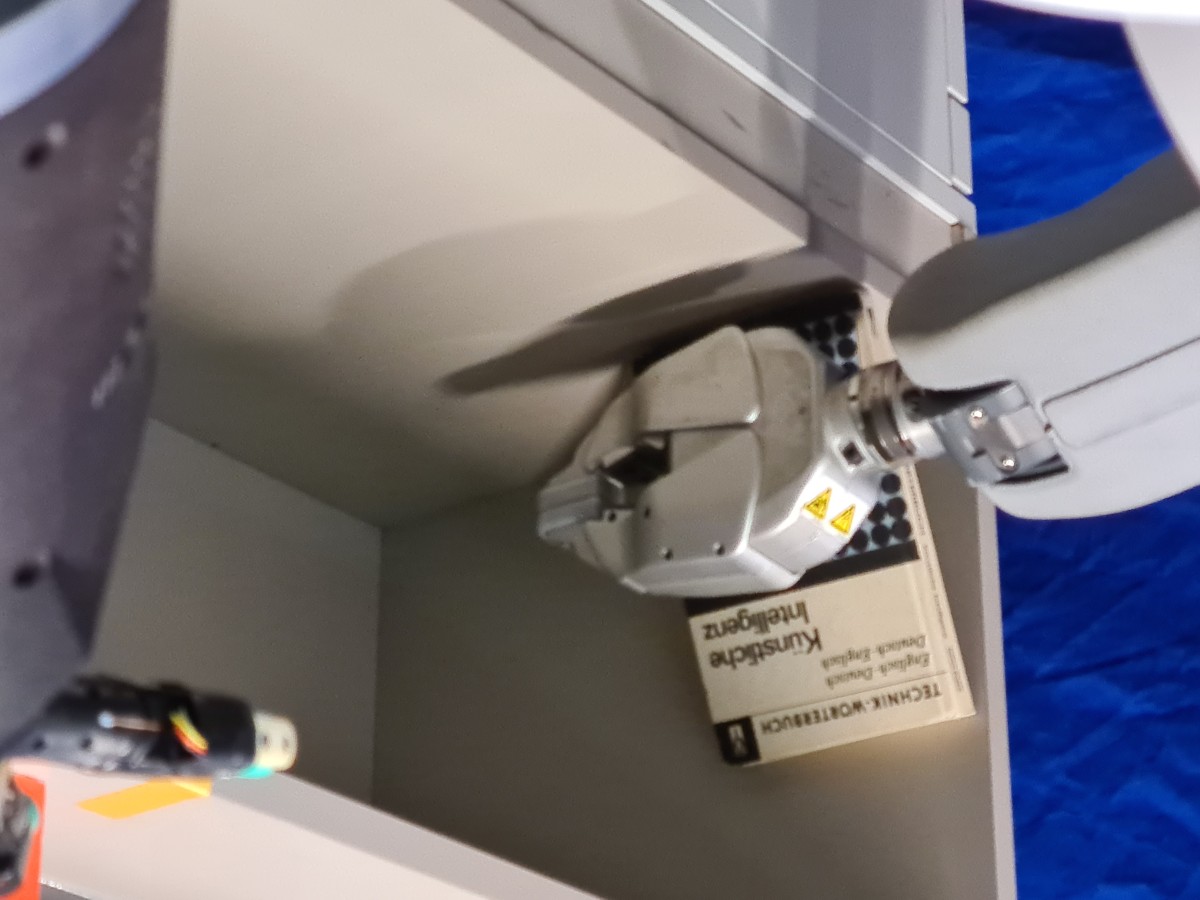}
\subcaption{Positioning next to the book to prepare pushing the book.}  
\end{subfigure}
\hfill
\begin{subfigure}[t]{.2\textwidth}
\centering
\includegraphics[angle=180, width=\linewidth]{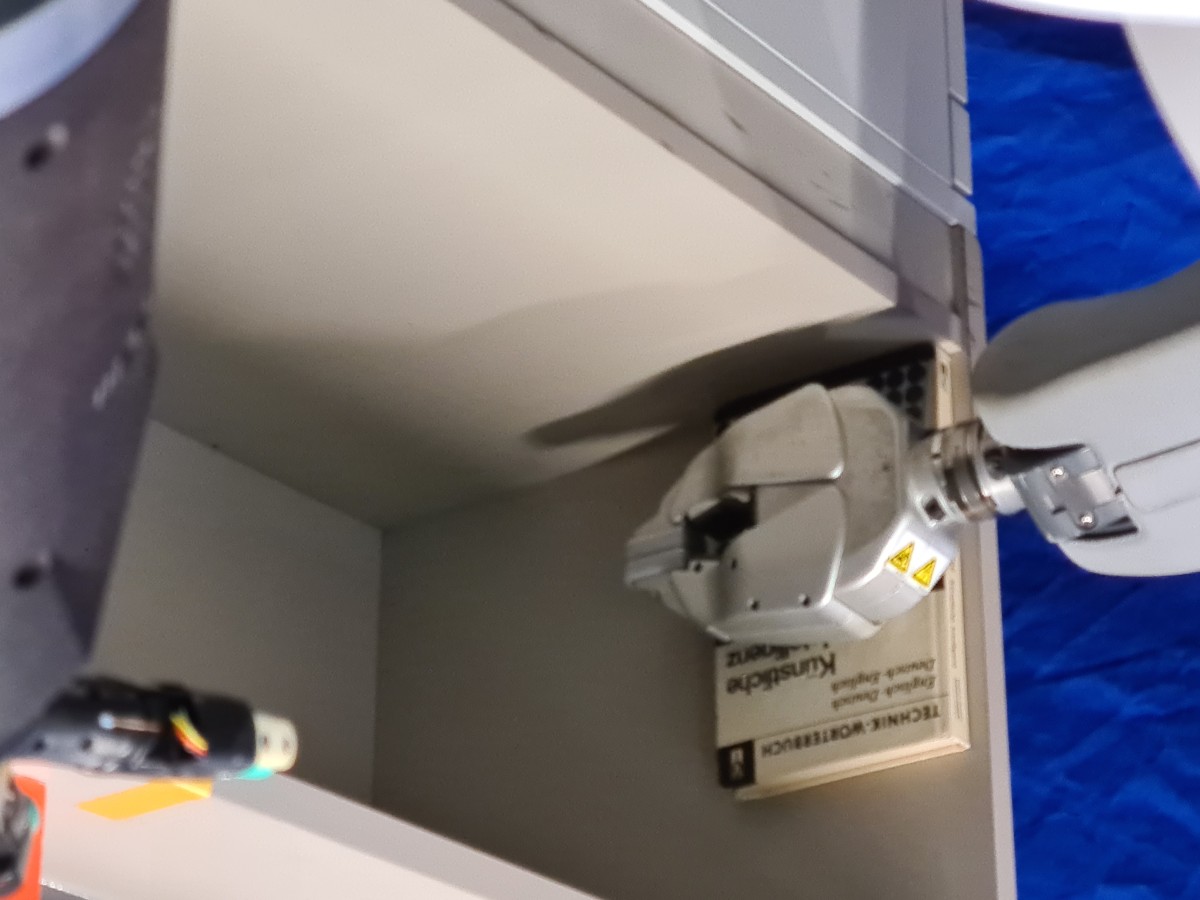}
\subcaption{Pushing the book against the shelf to let it stand upright.}  
\end{subfigure}
\caption{The process of placing the book.}
\label{fig:placing_process}
\end{figure}
For the placing procedure, we rely on the two finger gripper and its force-torque sensor. First, the book is moved next to the designated position and tilted slightly, so in the next steps it can be leaned against either the shelf wall or another book. This limits our implementation to not be able to place the book on its own or between other books. However, placing the book on its own can be easily implemented, if a way to stabilize the book during the placing can be found. Placing the book between other books, however, is similar to the famous peck insertion task, and solving this was outside the scope of our project. Therefore, we have decided not to try to implement any way to solve this problem.
The leaning against an object process is then done using the force-torque sensor. At the start, the book is moved as far as possible against the object. For the stopping point, a threshold for the change in force on the gripper between holding the book and touching the object is considered. Similarly, the next step is moving as far down as possible to touch the book to the shelf. Then, the gripper can then release the book, and the book is leaning against the object. Finally, the gripper can move again from the side against the book, pushing it upright against the object. This movement is also controlled using the force-torque sensor. Now, the book stands upright next to the designated object. The full process can be seen in \ref{fig:placing_process}.

While this process has proven its feasibility in our experiments, it also has its drawbacks. First, it requires a surface against which the books have to be leaned against, making it impossible to place the books in the middle of an empty shelf. However, when sorting books, the usual approach would be to place them from left to right anyway, making this limitation acceptable. The second, more limiting factor, is the problem, that placing a larger against a smaller book can result in failure, if the height difference during leaning the second book against the first is too great. This limits our process to only sort the books consistently by height, from highest to lowest. In the future, this issue might be able to be solved by implementing a different procedure, if the height difference is too great. One example would be to grasp the second book on the top of its spine with the two finger gripper and already placing it vertically. While this is not done, we have a working procedure for placing the picked books, with some limitations.

\section{Results (Shang-Ching Liu) }
\begin{figure}[ht]
    \centering
    \includegraphics[width=8cm]{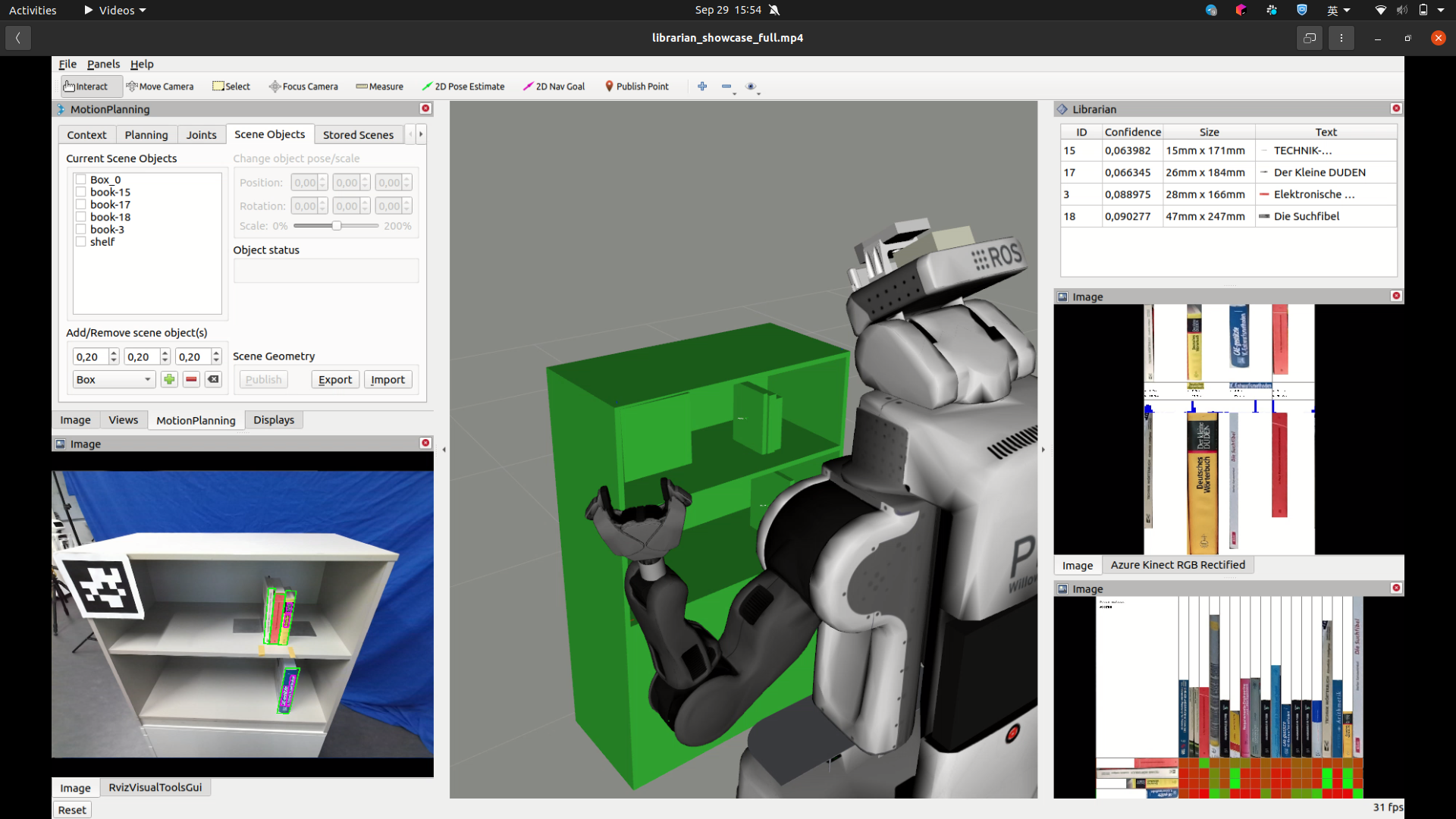}
    \caption{Demo result of RViz}
    \label{fig:result_rviz}
\end{figure}

The system has been organized into three modules, including librarian\_vision, librarian\_manipulation, and librarian\_common. In the final demo, we demonstrate sorting three books on the upper shelf and placing them on the lower shelf, as shown in figure \ref{fig:grasping_process}. You can see the book matching result easily in the RViz panel as figure \ref{fig:result_rviz}. After the robot recognizes the book, we turn off the perception pipeline and start the grasping task. The grasping performs cooperation from both hands and notices we have intentionally placed the front and back sides of the book in front of the camera vision focus area to be further processed for text reading or to eliminate the error in the perception area. Finally, place the sorted book on the lower shelf.

\section{Conclusion (Mykhailo / Björn)}
In this paper, we have successfully implemented a pipeline, which enables the robot to locate, pick and place different books within a bookshelf. By first using a bounding box approach to detect the book spines, we managed to locate the different books at previously unknown positions on the shelf. Through HSV matching, we then managed to identify the individual books, using the available database. Further, we utilized the database to find the correct order in which to grasp and sort the books, which is then accomplished by grasping them and placing them, sorted, in another area of the bookshelf. The grasping process utilized both different grippers from the PR-2 to be able to even pick a book from between other books. While our pipeline is not the most robust and has some limitations, our results show the feasibility of our approach and with additional work in the future, it should be possible to remove many of the current limitations and make the pipeline more robust. Some possible improvements for the vision part of the pipeline would include a larger training set for our YOLO network to prevent overfitting, making it more robust and allowing it to generalize better, or more fine-tuned metrics for our matching, making this part more robust as well. For the grasping part, having the robot move to be able to reach more areas of the shelf or modify the placing strategy as already described, to account for larger books getting placed against smaller ones would remove a lot of the current limitations. Afterward, the pipeline could be upgraded with more interaction between the different parts or new strategies for grasping, which would all be interesting improvements in the future.


\section*{Acknowledgment}
We would like to express our special thanks of gratitude to both supervisor Michael Görner and Dr. Norman Hendrich as well, who provide fully support to all issues during the project. Also, the research group Technical Aspects of Multimodal Systems (TAMS) giving rich experiment robot environment that make us came to know about so many new things.

\bibliographystyle{plain} 
\bibliography{refs} 

\section*{Appendix (Shang-Ching Liu)}
Following is the progress of the Trixi through developing:

\begin{figure}[ht]
\centering
\includegraphics[width=0.6\linewidth]{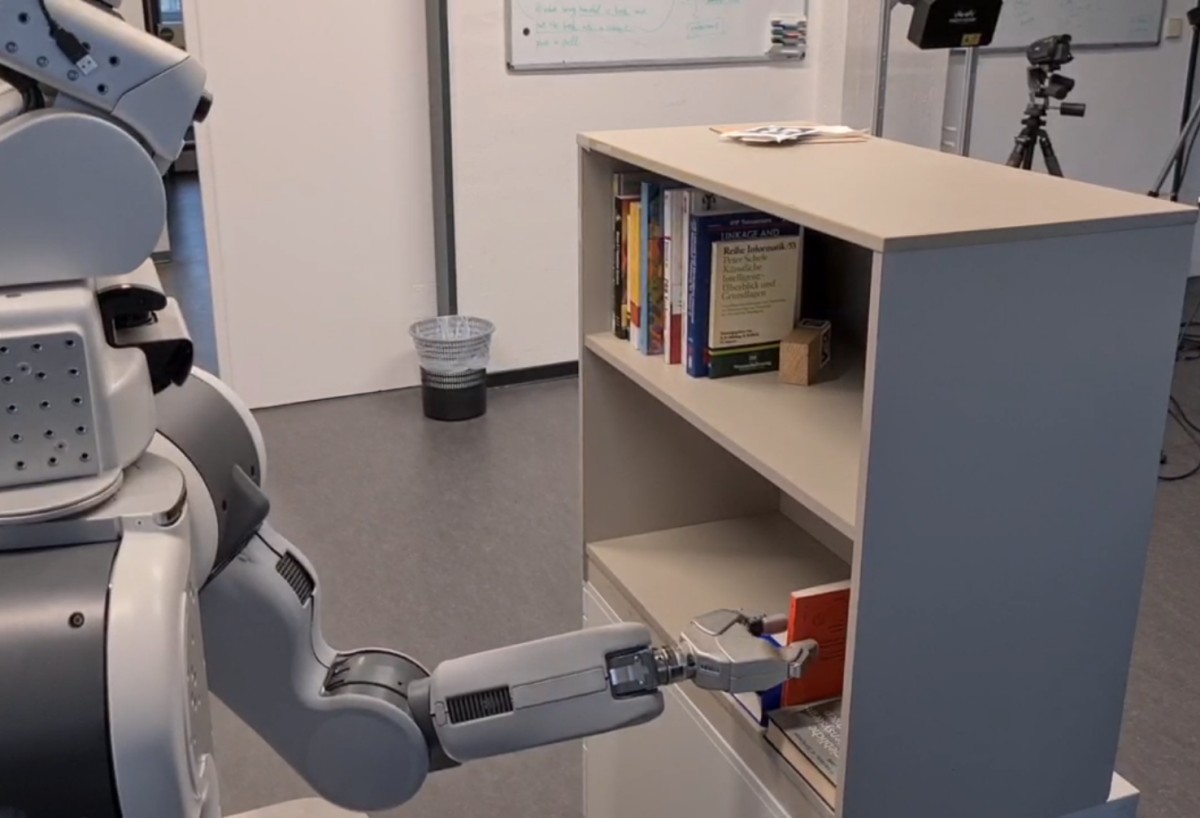}
\caption{Begin from one book being grasped by one gripper.}  
\end{figure}

\hfill

\begin{figure}[ht]
\centering
\includegraphics[width=0.4\linewidth]{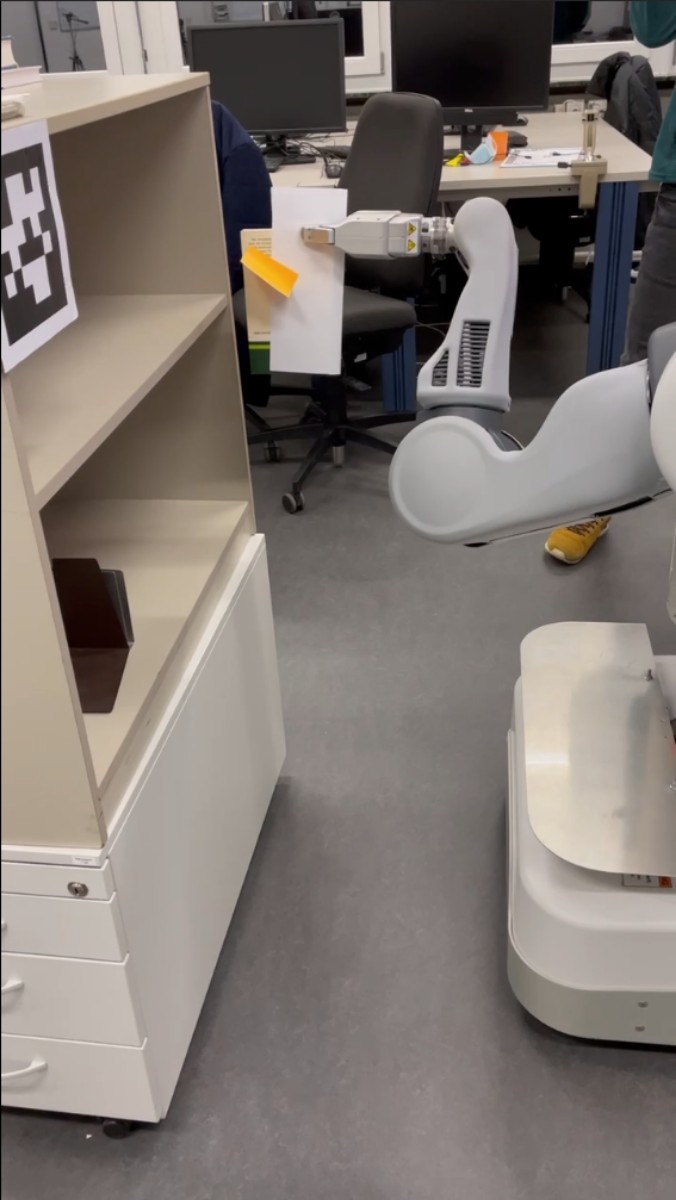}
\caption{Recognize the text and grasped specific book, however the text size matter a lot conflict with the book weight that robot gripper can offer, so we create a fake book cover here.}
\end{figure}

\hfill

\begin{figure}[ht]
\centering
\includegraphics[width=0.6\linewidth]{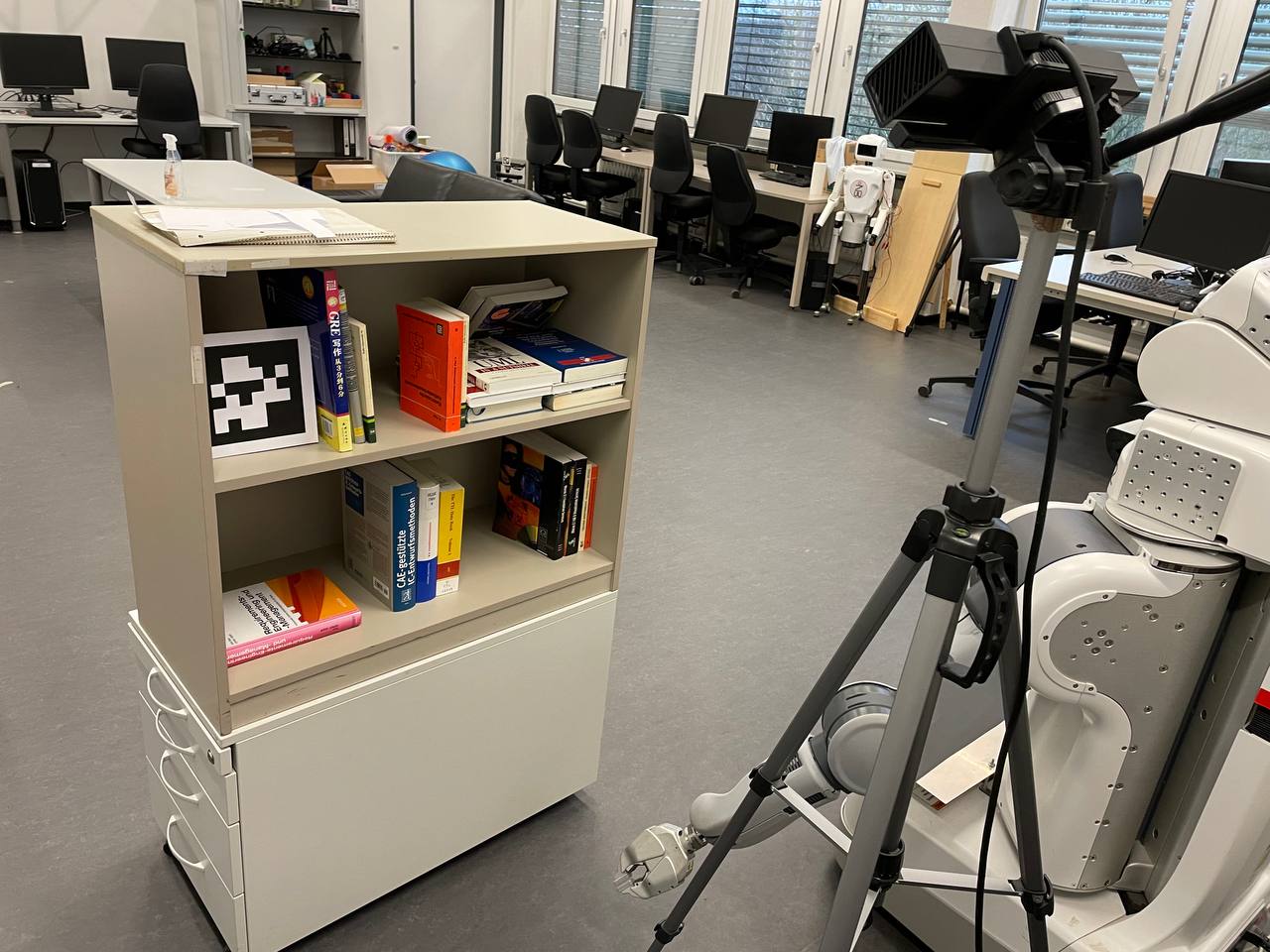}
\caption{Collecting the book spine database image.}  
\end{figure}

\hfill

\begin{figure}[ht]
\centering
\includegraphics[width=0.6\linewidth]{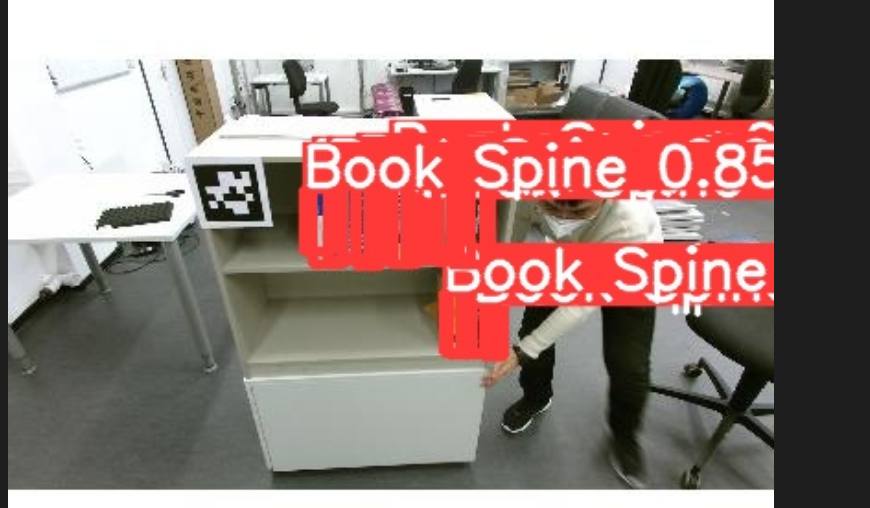}
\caption{Get quite good result from YOLOv5 to detect the book spine.}  
\end{figure}

\hfill

\begin{figure}[ht]
\centering
\includegraphics[width=0.6\linewidth]{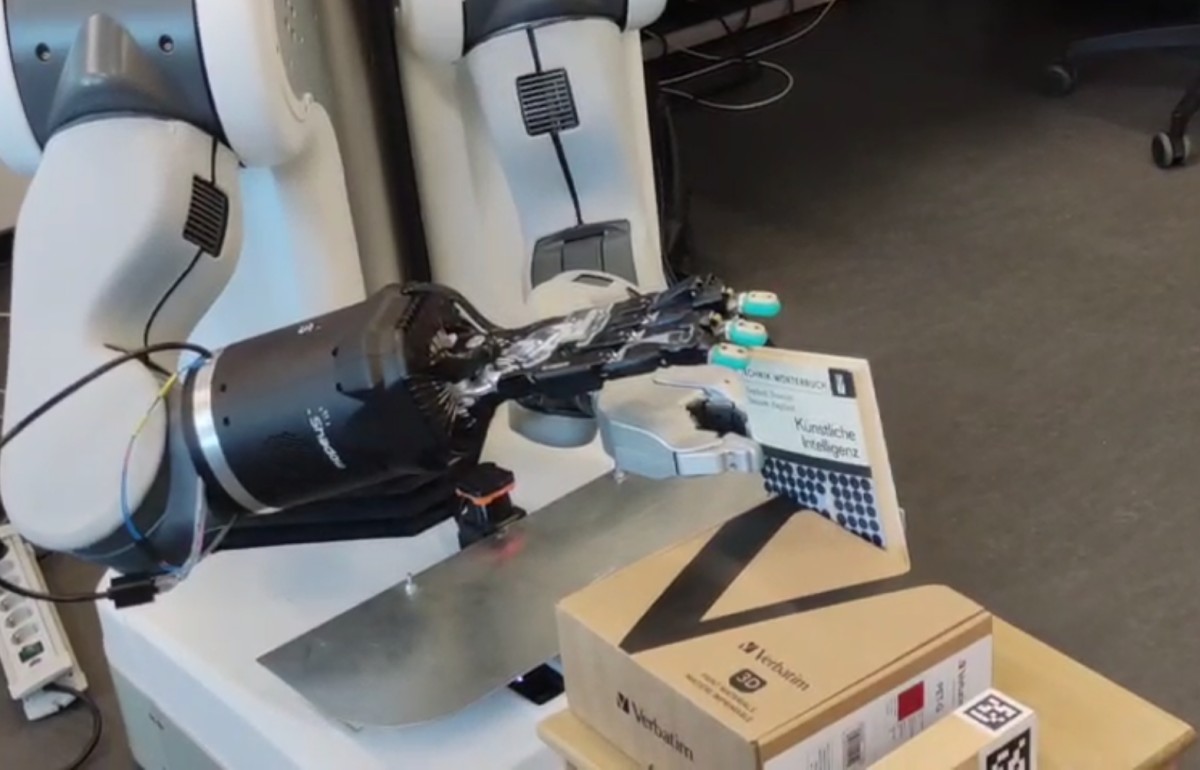}
\caption{First time pick the book by shadow-hand success.}  
\end{figure}

\hfill

\begin{figure}[ht]
\centering
\includegraphics[width=0.6\linewidth]{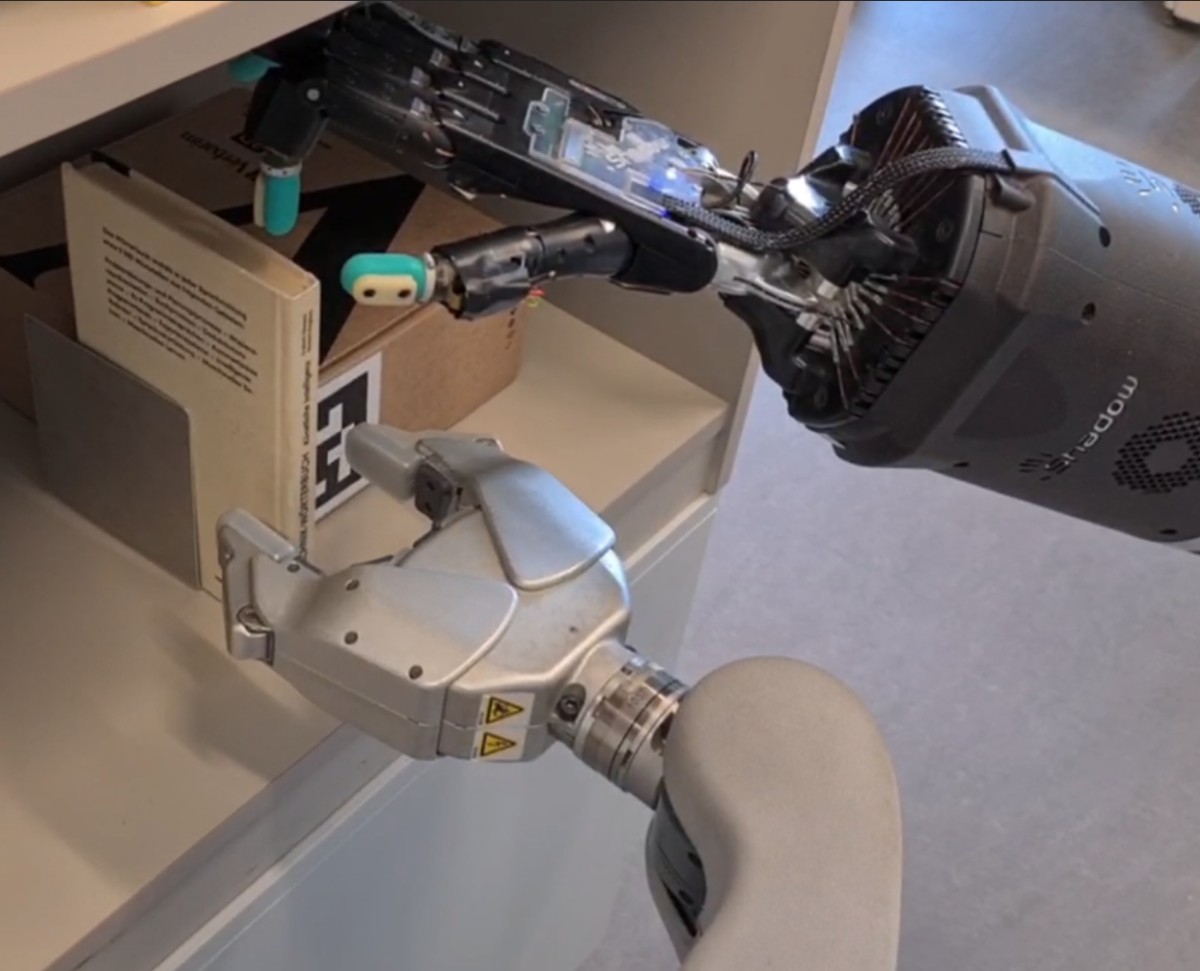}
\caption{First time, the cooperation between shadow hand and gripper success.}  
\end{figure}

\begin{figure}[ht]
\centering
\includegraphics[width=0.6\linewidth]{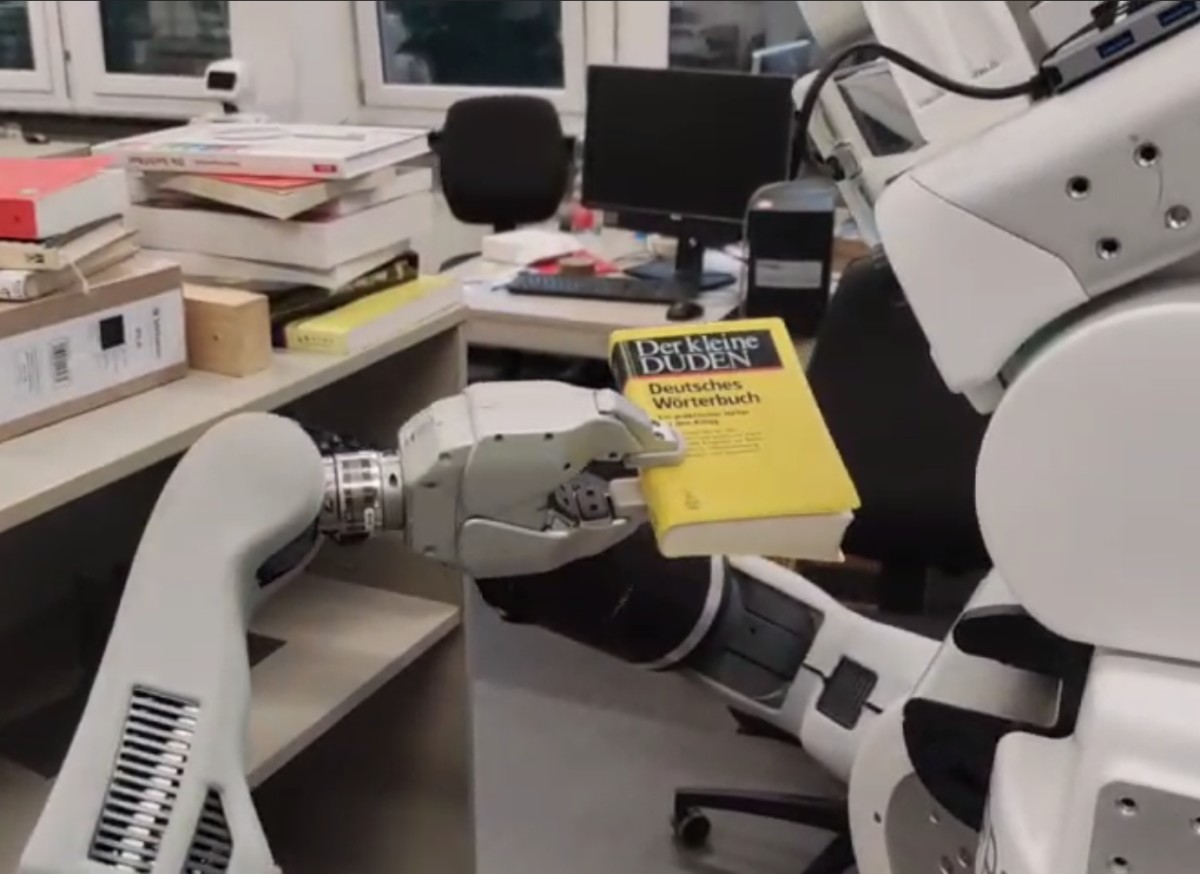}
\caption{First motion to read the book implement in Trixi.}  
\end{figure}

\end{document}